\documentclass{article}

\usepackage{microtype}
\usepackage{graphicx}
\usepackage{subfigure}
\usepackage{booktabs} %

\usepackage{hyperref}

\usepackage[accepted]{icml2024}

\usepackage{amsmath}
\usepackage{amssymb}
\usepackage{mathtools}
\usepackage{amsthm}

\usepackage[capitalize,noabbrev]{cleveref}

\theoremstyle{plain}
\newtheorem{theorem}{Theorem}[section]
\newtheorem{proposition}[theorem]{Proposition}

\theoremstyle{definition}
\newtheorem{definition}[theorem]{Definition}

\theoremstyle{remark}

\usepackage[textsize=tiny]{todonotes}

\usepackage{subcaption}
\usepackage[mathscr]{eucal}
\usepackage[inline]{enumitem}
\usepackage{wrapfig}
\usepackage{framed}
\usepackage{caption}
\usepackage{multirow}

\usepackage{stmaryrd}
\usepackage{dsfont}

\icmltitlerunning{Multi-Marginal Matching Gap}

\usepackage{listings}
\usepackage{amsmath,amsfonts,bm,amsthm}
\usepackage{pifont}
\usepackage{graphicx}
\usepackage{wrapfig}
\usepackage{xcolor}
\usepackage{sidecap}

\newcommand{\MODELshort}{{\rm{M3G}}}

\definecolor{maroon}{rgb}{0.5, 0.0, 0.0}

\def\eqref#1{Eq.~(\ref{#1})}

\def\appref#1{\S\ref{#1}}

\def\1{\bm{1}}

\DeclareMathAlphabet{\mathsfit}{\encodingdefault}{\sfdefault}{m}{sl}
\SetMathAlphabet{\mathsfit}{bold}{\encodingdefault}{\sfdefault}{bx}{n}

\newcommand{\R}{\mathbb{R}}

\newcommand{\mfg}{\mathrm{M3G}}

\DeclareMathOperator*{\argmin}{arg\,min}

\def\*#1{\mathbf{#1}}

\def\obj{h}
\def\OT{\mathrm{OT}}

\begin{document}

\twocolumn[
\icmltitle{Contrasting Multiple Representations with the Multi-Marginal Matching Gap}

\icmlsetsymbol{equal}{*}

\begin{icmlauthorlist}
\icmlauthor{Zoe Piran}{equal,app,huj}
\icmlauthor{Michal Klein}{equal,app}
\icmlauthor{James Thornton}{app}
\icmlauthor{Marco Cuturi}{app}
\end{icmlauthorlist}

\icmlaffiliation{app}{Apple}
\icmlaffiliation{huj}{Hebrew University Jerusalem}

\icmlcorrespondingauthor{Marco Cuturi}{cuturi@apple.com}

\icmlkeywords{Machine Learning, ICML, Optimal Transport, SSL}

\vskip 0.3in
]

\printAffiliationsAndNotice{\icmlEqualContribution} %

\begin{abstract}
Learning meaningful representations of complex objects that can be seen through multiple ($k\geq 3$) views or modalities is a core task in machine learning.
Existing methods use losses originally intended for paired views, and extend them to $k$ views, either by instantiating $\tfrac12k(k-1)$ loss-pairs, or by using reduced embeddings, following a \textit{one vs. average-of-rest} strategy. 
We propose the multi-marginal matching gap ($\mfg$), a loss that borrows tools from multi-marginal optimal transport (MM-OT) theory to simultaneously incorporate all $k$ views.
Given a batch of $n$ points, each seen as a $k$-tuple of views subsequently transformed into $k$ embeddings, our loss contrasts the cost of matching these $n$ ground-truth $k$-tuples with the MM-OT polymatching cost, which seeks $n$ optimally arranged $k$-tuples chosen within these $n\times k$ vectors.
While the exponential complexity $O(n^k$) of the MM-OT problem may seem daunting, we show in experiments that a suitable generalization of the Sinkhorn algorithm for that problem can scale to, e.g., $k=3\sim 6$ views using mini-batches of size $64~\sim128$.
Our experiments demonstrate improved performance over multiview extensions of pairwise losses,  for both self-supervised and multimodal tasks.
\end{abstract}

\section{Introduction}\label{sec:intro}
Learning meaningful representations of complex objects that can be seen through multiple views or modalities is a core task in machine learning. These representations may be trained separately for each modality, as a preliminary step towards zero-shot learning~\citep{NIPS2009_1543843a,NIPS2013_2d6cc4b2,frome2013devise}. In that scenario, modalities can be heterogeneous, as with images and text~\citep{radford2021learning,schuhmann2022laion}, or beyond~\citep{deldari2022beyond}; or homogeneous, e.g. various channels of the same timeseries~\citep{khaertdinov2021contrastive,cheng2020subject,wen2020time,brusch2023multi,banville2021uncovering,tonekaboni2021unsupervised,kiyasseh2021clocs}. In the closely related task of self-supervised learning (SSL), a single embedding backbone may be considered instead, and applied to multiple views/augmentations of the same object~\citep{chen2020simple,caron2020unsupervised,bardes2022vicregl,assran2023self,tsai2021selfsupervised}. %

\paragraph{Learning with Pairs.} Whether applied to multiview or multimodal learning, these approaches were originally proposed for $k=2$ different representations (e.g. arising from two modalities or two augmentations). Most of them rely on contrastive learning~\citep{gutmann2010noise,oord2018representation} as a blueprint, using, for instance, the InfoNCE loss. The InfoNCE loss promotes encoders that produce nearby representations for \textit{two} inputs that arise from the same object (either with different views or modalities), and far-away representations for any other pair. Alternatively, BYOL~\citep{grill2020bootstrap} uses only positive pairs, and relies instead on a pair of encoders with tied parameters.
\begin{figure*}[ht] 
\includegraphics[width=\textwidth]{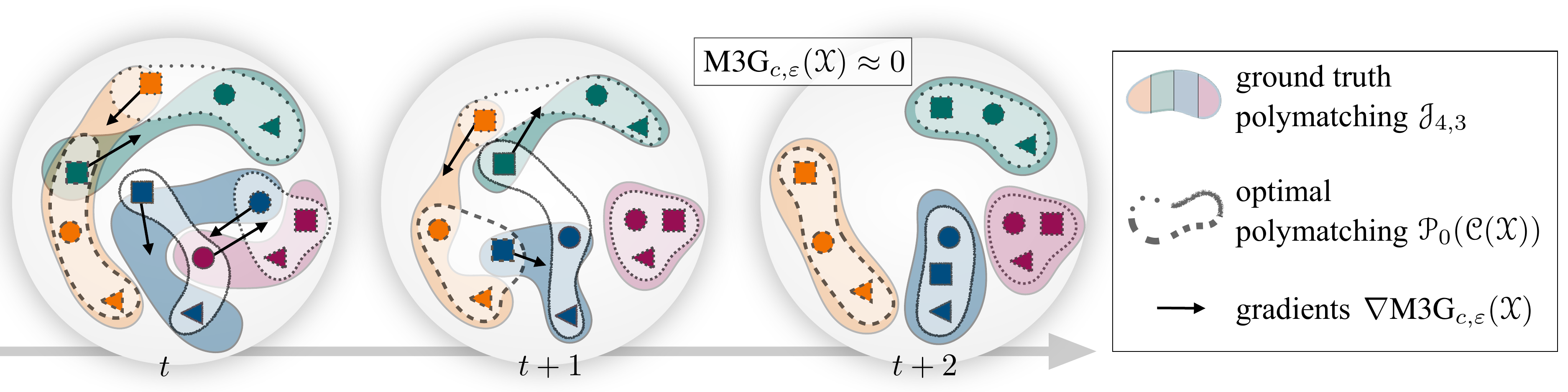}
\caption{\textit{(left)} Embeddings for $n=4$ points (identified using 4 colors), each given in $k=3$ views (differentiated using 3 shapes) in $d=2$ dimensions. The ground-truth polymatching of these points is known: to each color its 3 shapes, as illustrated with colored cliques, and described mathematically as a tensor $\mathscr{J}_{4,3}$. Their initial configuration in space indicates, assuming one solves a multi-marginal optimal transport problem parameterized with the cost tensor $\mathscr{C}(\mathscr{X})$, a \textit{different} polymatching $\mathscr{P}_0(\mathscr{C}(\mathscr{X}))$. That difference (quantified as a difference in their \textit{matching objectives}) defines the $\mfg$ loss (see Def.~\ref{def:m3g} for a precise definition of what $c,\varepsilon$ refer to). A high $\mfg$ indicates, as shown on the left, a large discrepancy between the ground-truth matching's cost and that of the optimal polymatching. This loss will gradually displace points so that, ideally, upon convergence and after consecutive updates (visualized in \textit{(middle)} and \textit{(right)} plots), both ground-truth and optimal polymatchings coincide in their objective. For additional intuition see Animation \href{https://mlrapp.github.io/m3g/}{1}, presenting the gradient flow of $\mfg$ over a toy problem.}\label{fig:intro}
\end{figure*}

\paragraph{Learning with $k\geq 3$ Representations.}
As representation learning eyes more ambitious tasks, practitioners are tempted to incorporate more than two views/modalities~\citep{alayrac2020self,akbari2021vatt,girdhar2023imagebind}. Various strategies have been proposed to handle $k\geq 3$ representations that can cope with the limitation of pairwise losses~\citep{NEURIPS2019_ddf35421,tian2020contrastive,tsai2020self}. For instance, one may handle $k$ representations by averaging all possible $\tfrac12k(k-1)$ pairwise losses~\citep{NEURIPS2019_ddf35421,tian2020contrastive}; Alternatively, one may average embeddings~\citep{pototzky2022fastsiam}, effectively comparing each of the $k$ representations to the average of the remaining $k-1$ embeddings. None of these approaches do leverage, however, the knowledge that these $k$ representations should be \textit{simultaneously} coherent, by looking at $k$-tuple of points rather than $\tfrac12k(k-1)$ pairs.

\paragraph{Our Contributions.} We propose a novel approach that fully leverages the ground-truth knowledge that a single input data point should be viewed, holistically, as $k$-tuples of embeddings. Our contrastive loss is tailored for multiple ($k \geq 3$) views, without a reduction to pairwise comparisons. This global view is provided by solving polymatching problems using entropy-regularized multi-marginal optimal transport (MM-OT). More precisely:
\begin{itemize}[leftmargin=.3cm,itemsep=.0cm,topsep=0cm,parsep=2pt]
    \item After providing background on SSL and MM-OT in \S\,\ref{sec:back}, we present in \S~\ref{sec:methods} the $\mfg$ loss to measure the contrast of a configuration of an $n$-batch of $k$-tuples of points. $\mfg$ subtracts the lowest matching cost achieved by an MM-OT solver (MM-\citeauthor{sinkhorn1964relationship}) to the matching cost of the ground-truth identity matching tensor available to the user. We study computational and theoretical properties of $\mfg$, highlight the freedom to choose any multiway cost function defined on $k$-tuple of points, and show how to use $\mfg$ for representation learning. 
    \item We provide experimental evidence in \S~\ref{sec:experiments} that the $\mfg$ loss improves on extensions of pairwise losses in a variety of self-supervised and multimodal tasks, using the ImageNet-1k dataset~\citep{deng2009imagenet}, DomainNet~\citep{he2020momentum}) and time-series electroencephalography (EEG) data from PhysioNet~\citep{goldberger2000physiobank, ghassemi2018you, kemp2000analysis}.
\end{itemize}

\textbf{Notation.} We use bold fonts for vectors $\*x,\*y,\dots$ in $\mathbb{R}^d$ and matrices $\*X,\*Y, \*P,\*C\dots$ in $\mathbb{R}^{n\times d}$ or $\mathbb{R}^{n\times n}$; curved fonts $\mathscr{X},\mathscr{Y}, \mathscr{P}, \mathscr{C}\dots$ for tensors of dimension $3$ and more. For an integer $k$, we set $\llbracket k \rrbracket:=(1,\dots, k)$, and for two integers $\ell < m$, $\llbracket \ell, m \rrbracket := (\ell, \ell+1, \dots, m)$. 

\section{Background: SSL and MM-OT}\label{sec:back}

\subsection{Joint embeddings with student-teacher architecture}\label{sec:back_student_teacher}

We rely in this work on joint embedding student-teacher architectures~\citep{balestriero2023cookbook,grill2020bootstrap}.
This setting consists of a tied pair of \textit{online}  (student) and \textit{target} (teacher) networks. The student network contains three components--an encoder, a projector and a predictor. The teacher is based on the online network, omitting the predictor head.   
Importantly, the parameters of the latter are updated using an exponential moving average (EMA) of the student's parameters.
Setting, for instance, the index $s$ to be the student, and $t$ the teacher, the parameters $\theta_t$ are simply updated as $\theta_{t} \leftarrow (1-\rho)\theta_{s}+ \rho \theta_{s}$ after each $\theta_s$ update, with EMA parameter $0<\rho<1$.

\subsection{Learning embeddings with pairwise losses}\label{sec:back_infonce}

\textbf{Learning with $k=2$ views.}
In a standard SSL setup, one selects a batch of $n$ items $(z_i)_i:=(z_1, \cdots, z_n)$ alongside two augmentation pipelines $\mathcal{A}_1$ and $\mathcal{A}_2$; applies both augmentations to each item in the batch, yielding a list of $n$ pairs of objects, $(\mathcal{A}_1(z_{i}),\mathcal{A}_2(z_{i}))_i$. These are then passed through parameterized neural networks, $f_{\theta_1}, g_{\theta_2}$,  that produce vector representations $\*x^1_{i} := f_{\theta_1}(\mathcal{A}_1(z_{i}))$ and $\*x^2_{i} := g_{\theta_2}(\mathcal{A}_2(z_{i}))$.  This results in $\*X^1 := (\*x^{1}_i)_i, \*X^2 = (\*x^{2}_i)_i$, two $n\times d$ matrices of embeddings, which we assume throughout the paper to lie on the $d$-sphere, i.e. their norms are equal to 1. 
These views are then fed to a pairwise loss, $\mathcal{L}_{\rm{pair}}\left(\*X^{1},\*X^{2}\right)$ used to fit either or both parameters $\theta_1, \theta_2$.
 The seminal approach of SimCLR~\citep{chen2020simple} considered a variant of the \rm{InfoNCE} loss~\citep{oord2018representation,gutmann2010noise} defined as:

\begin{equation}\label{eq:infoNCE}
\mathcal{L}_{\rm{InfoNCE}}(\*X^{1},\*X^{2}) = 
   - \frac{1}{n}\sum_{i=1}^n \log \!\left(\frac{e^{\tfrac{\langle\*x^{1}_{i},\*x^{2}_{i}\rangle}{\tau}}}
   {\sum_{j} e^{\tfrac{\langle\*x^{1}_{i},\*x^{2}_{j}\rangle}{\tau}}}\!\right).
\end{equation}

\citet{grill2020bootstrap} propose an alternative that leverages an assymetric student-teacher setting (\S~\ref{sec:back_student_teacher}), to focus exclusively on paired positive examples: 

\begin{equation}\label{eq:byol}
\mathcal{L}_{\rm{BYOL}}(\*X^{1},\*X^{2}) = 
   2 - \frac{2}{n}\sum_{i=1}^n\langle\*x^{1}_{i},\*x^{2}_{i}\rangle.
\end{equation}

\textbf{Extending pairwise losses to $k\geq 3$ views.} Recent contrastive approaches rely on $k \geq 3$ views to improve model performance~\citep{caron2020unsupervised, tian2020contrastive, zhou2022mugs, bardes2022vicreg}. For example, a multi-crop strategy for images adds various views at
different resolutions, rather than two full-resolution views~\citep{caron2020unsupervised}, to extract more information from a single input object~\citep{balestriero2023cookbook,hoffer2020augment}. The number of samples the model sees is effectively increased at each batch to $n\times k$ instances, $k$ views for a batch of $n$ images. Each of these instances is then represented as a $d$-dimensional vector, all of which can be stored as a 3D tensor: $k$-views for $n$ points in $d$-dimensions, $\mathscr{X}\in\mathbb{R}^{k\times n\times d}$, $\mathscr{X}=[\*X^1,\dots,\*X^k]$ where each $\*X^l$ gathers the $n$ objects as seen from the $\ell$-th view, namely an $n\times d$ matrix $\*X^\ell = [ \*x_1^\ell, \dots, \*x_n^\ell]$.
To handle multiple views, a \textit{pairwise} contrastive loss $\mathcal{L}_{\rm{pwe}}$ can be defined by aggregating all possible pairs, $\tfrac12k(k-1)$ in total, of $\mathcal{L}_{\rm{pair}}$ losses,
\begin{equation}\label{eq:sum}
\mathcal{L}_{\rm{pwe}}(\mathscr{X}) = 
   \frac{2}{k(k-1)} \sum_{\ell< m}^{k} \mathcal{L}_{\rm{pair}}\left(\*X^{\ell},\*X^{m} \right)\,,
\end{equation}
The summation can be performed on all, a subset of the pairs, e.g. restricting $\ell$ to sweep ${1,2}$ and taking $m \in \llbracket k \rrbracket$ ~\citep{caron2021emerging,grill2020bootstrap}, or with a different aggregation method~\citep{shidani2024polyview}. An alternative way to fall back on using a parwiese loss, is averaging representations beforehand, and applying the loss in a \textit{one vs. average-of-rest} fashion. That is, for each view $\ell$, the embeddings $\*X^{\ell}$, are compared to the \textit{average} of all remaining views, $\bar{\*X}^{-\ell} := \tfrac1{k-1}\sum_{m\neq \ell} \*X^{m}$ as presented in~\citep{pototzky2022fastsiam,liang2024multiple}, defining the loss, 
\begin{equation}\label{eq:mvp}
\mathcal{L}_{\rm{ave}}(\mathscr{X}) = 
   \frac{1}{k} \sum_{\ell=1}^{k} \mathcal{L}_{\rm{pair}}\left(\*X^{\ell},\bar{\*X}^{-\ell} \right)\,.
\end{equation}
These two approaches only look at the entire representation tensor $\mathscr{X}$ two slices at a time, either by comparing $\*X^\ell$ to another $\*X^m$, $\bar{\*X}^{-\ell}$, or $\*X^\ell$ to a combination of the other $(\*X^{m})_{m\ne \ell}$. MM-OT, introduced next, will serve as the workhorse to provide the first holistic loss for multiple views, leveraging the entire distribution described in $\mathscr{X}$.

\subsection{Multi-Marginal Optimal Transport (MM-OT)}
We borrow notations from~\citep[Chap. 4]{Peyre2019computational}, restricting our attention to \textit{matching} problems (uniform marginals of the same size). As a warm-up to the multi-marginal case, we start with two marginals.

\textbf{Regularized Bistochastic Matching.}
 Consider a cost matrix $\*C\in\mathbb{R}^{n\times n}$.%
The entropy regularized matching cost of $\*C$, parameterized by regularization $\varepsilon\ge 0$, is the output of the following minimization:

\begin{equation}\label{eq:regwass}
      \OT_{2,\varepsilon}(\*C)= \min_{\*P\in \mathcal{B}_{n,2}} \langle \*P, \*C\rangle +\varepsilon \langle \*P, \log \*P -1\rangle\,,
\end{equation}
where $\mathcal{B}_{n,2}$ is the \citeauthor{birkhoff} polytope of bistochastic \textit{matrices},
\begin{equation}\label{eq:birk}
\mathcal{B}_{n,2} := \{\*P\in\mathbb{R}^{n\times n}_+\, |\, \*P\mathds{1}_{n}=\*P^T\mathds{1}_{n} =\mathds{1}_{n}/{n}\}\,.
\end{equation}
For $\varepsilon=0$, one recovers the optimal assignment problem, used for instance to compute a loss between lists of annotations in an image~\citep{carion2020end}. When $\varepsilon>0$, the problem can be solved with the \citeauthor{sinkhorn1964relationship} fixed point iterations, with faster execution on accelerators~\citep{cuturi2013sinkhorn}.%

\textbf{Regularized Polystochastic Matchings.}
We consider the generalization to multidimensional cost tensors of the matching problem, moving away from the bipartite setting described above. Such problems arise when comparing $k\geq 3$ families of points simultaneously, to solve \textit{polypartite} matching problems. The MM-OT~\citep{gangbo1998optimal,pass2015multi} problem and its entropic regularization~\citep{benamou2015iterative} generalize~\eqref{eq:regwass} by searching for $k$-polymatchings, represented with their relaxation as polystochastic tensors~\citep{benson2014counting}. To introduce these approaches, we need a few more notations.

\textbf{Polystochastic Tensors.}
We consider the set of $k$-dimensional tensors, of size $n$ for each slice:
$$\mathcal{T}_{n,k}:=\mathbb{R}^{\tiny{\overbrace{n\times\dots\times n}^{k\, \text{times}}}}.$$
Let $\mathds{1}_{n,k}$ be the tensor in $\mathcal{T}_{n,k}$ containing ones, including $\mathds{1}_n:= \mathds{1}_{n,1}$ the $n$-vector of ones. For a tensor $\mathscr{P}\in\mathcal{T}_{n,k}$, and $\ell\leq k$, we write $\textrm{m}_\ell$ for the contraction of the tensor along all of its slices expect for the $\ell$-th one. Using the tensordot operator (with 1-indexing, not 0 as used by default in python), this is equivalent to, writing $I_\ell = (\llbracket k \rrbracket \setminus \ell, \llbracket k-1 \rrbracket)$ for the pair of contraction indices,
$$
\textrm{m}_\ell(\mathscr{P}) = \texttt{tensordot}(\mathscr{P},\mathds{1}_{n,k-1},{I_\ell}) \in \mathbb{R}^n\,.
$$

We define, by analogy to~\eqref{eq:birk}, the polytope of $k$-polystochastic tensors,
\begin{equation}\label{eq:birkn}
\mathcal{B}_{n,k} := \{\mathscr{P}\in\mathcal{T}_{n,k} \,|\, \forall \ell\leq k,\, \textrm{m}_\ell(\mathscr{P}) = \mathds{1}_n/n\}\,.
\end{equation}
We can now generalize~\eqref{eq:regwass} by replacing the suffix $2$ by $k$ in these expressions, to define, for any cost tensor $\mathscr{C}\in\mathcal{T}_{n,k}$,
\begin{align}
      \OT_{k,\varepsilon}(\mathscr{C})&:= \min_{\mathscr{P}\in \mathcal{B}_{n,k}} \obj_{k,\varepsilon}(\mathscr{P}, \mathscr{C}),\,\text{where} \nonumber\\ %
      \obj_{k,\varepsilon}(\mathscr{P}, \mathscr{C})&:=\langle \mathscr{P}, \mathscr{C}\rangle +\varepsilon \langle \mathscr{P}, \log \mathscr{P} -1\rangle\,.\label{eq:regwassk}
\end{align} 

\textbf{Dual Formulation.}
\eqref{eq:regwassk} admits an unconstrained dual formulation, using a few more notations:
For an $n\times k$ matrix $\*F$, stored as $k$ column vectors of size $n$, $\*F = (\*f^1,\dots, \*f^k)\in\mathbb{R}^{n\times k}$, we define the tensor sum operator, which to a matrix in $\mathbb{R}^{n\times k}$ associates a tensor in $\mathcal{T}_{n,k}$ as follows,
$$
\bigoplus \*F := \*f^1 \oplus \dots \oplus\*f^k,\, \text{i.e.} \left[\bigoplus \*F\right]_{i_1\dots i_k}\!\!\!\!\!\!\!\! = \*f^1_{i_1} + \dots + \*f^k_{i_k}.
$$
where all indices $1\leq i_1, \dots, i_k\leq n$. In that case, one has the following equivalence with~\eqref{eq:regwassk},
\begin{equation}\label{eq:dualwass}\OT_{k,\varepsilon}(\mathscr{C}) = 
\max_{\*F} \tfrac1n\mathds{1}_n^T \*F \mathds{1}_k - \varepsilon \langle e^{\frac{\bigoplus \*F-\mathscr{C}}{\varepsilon}}, \mathds{1}_{n,k}\rangle
\end{equation}

and the following primal-dual relationship among the optimal solutions $\mathscr{P}^\star$ of~\eqref{eq:regwassk} and $\*F^\star$ of~\eqref{eq:dualwass}:
\begin{equation}\label{eq:solregwassk_fact}
\mathscr{P}^\star = \exp\left(\left(\bigoplus \*F^\star-\mathscr{C}\right)/\varepsilon\right)\,.
\end{equation}

\textbf{Multi-Marginal \citeauthor{sinkhorn1964relationship}.}~\eqref{eq:dualwass} can be solved using the multi-marginal \citeauthor{sinkhorn1964relationship} (MM-S) algorithm described in Alg.~\ref{alg:mms}. This algorithm outputs, using~\eqref{eq:solregwassk_fact}, the optimal polystochastic tensor associated to $\mathscr{C}$:
\begin{equation}\label{eq:solregwassk}
\mathscr{P}_\varepsilon(\mathscr{C}):= \argmin_{\mathscr{P}\in \mathcal{B}_{n,k}} \obj_{k,\varepsilon}(\mathscr{P}, \mathscr{C}),
\end{equation}
 Note that Alg.~\ref{alg:mms} requires introducing the log-sum-exp operator: For a tensor $\mathscr{A}\in\mathcal{T}_{n,k}$, and a subset $I\subset \llbracket k \rrbracket$ of slices, $\textrm{LSE}(\mathscr{A}, I)$ denotes the log-sum-exp operator on such slices (this corresponds to $\texttt{log sum}(\exp(\mathscr{A}), \texttt{axis}=I)$, with a $1$ indexing convention).

The theoretical complexity of MM-S \citep{lin2022complexity} is $\mathcal{O}(k^3n^k\varepsilon^{-2})$, and involves in practice $k$ tensor reductions at each step, as highlighted in the for loop of Alg.~\ref{alg:mms}. As with the standard \citeauthor{sinkhorn1964relationship} algorithm, smaller $\varepsilon$ requires a larger number of iterations, and early stopping can be controlled with the tolerance parameter $\alpha$. In our experiments, we set $\alpha=10^{-3}$ and study the impact of $\varepsilon$ on the number of iterations needed to converge in \S\ref{tab:cost_ablation}.

\begin{algorithm}[tb]
\caption{Multi-marginal \citeauthor{sinkhorn1964relationship} (MM-S)}\label{alg:mms}
\begin{algorithmic}
   \STATE {\bfseries input:} cost tensor $\mathscr{C}\in\mathcal{T}_{n,k}$, regularization $\varepsilon$, tol. $\alpha$.
   \STATE $\*F = [\*f^1, \dots, \*f^k] = \mathbf{0}_{n\times k}$
   \REPEAT
   \FOR{$\ell, 1\leq \ell\leq k$}
        \STATE {$\*f^\ell \leftarrow -\varepsilon \left(\textrm{LSE}\left(\frac{\bigoplus \*F-\mathscr{C}}{\varepsilon}, \llbracket k \rrbracket \setminus \ell\right) +\log n\right)$}
   \ENDFOR
   \STATE {$\mathscr{P}\leftarrow\,\exp\left(\left(\bigoplus \*F-\mathscr{C}\right)/\varepsilon\right)$,\\
   $\delta \leftarrow\, \sum_{\ell=1}^{k} \|\textrm{m}_\ell(\mathscr{P})-\frac{\mathds{1}_n}{n}\|_1$}
   \UNTIL{$\delta<\alpha$}
   \STATE {\bfseries output:}\\ \quad Polystochastic tensor $\mathscr{P}\in\mathcal{T}_{n,k}$\,,\\ \quad MM-OT cost $\OT_{k,\varepsilon}(\mathscr{C})= \tfrac{1}{n}\langle \*F, \mathds{1}_{n\times k} \rangle - \varepsilon \langle \mathscr{P}, \mathds{1}_{n,k}\rangle$\,.
\end{algorithmic}
\end{algorithm}

\section{Multi-Marginal Matching Gap (\MODELshort)}\label{sec:methods}
We present our main contribution, the multi-marginal matching gap ($\mfg$) loss. The loss takes a $k\times n \times d$ tensor $\mathscr{X}$ of $k$ views for $n$ points of a batch, all represented as $d$-dimensional vectors. As sketched in Figure~\ref{fig:intro}, the loss quantifies, informally speaking, whether the $k$ views for each of the $n$ points cluster sufficiently when taken as a whole, relative to all other points. 

\subsection{Ground-Truth and Multiway Costs}

We introduce two crucial elements needed to define the $\mfg$: the ground-truth polymatching provided by batches of aligned points, and a cost function that quantifies the concentration of a $k$-tuple of vectors.

\textbf{Ground-Truth Polymatching. } Let $\mathscr{J}_{n,k}$ be the identity tensor in $\mathcal{T}_{n,k}$ divided by $n$. This is the tensor of zeros, except for the $n$ diagonal indices, which are all equal to $\tfrac{1}{n}$: 
$$[\mathscr{J}_{n,k}]_{i_1,\dots,i_n}= \tfrac1n \mathbf{1}_{i_1=\dots=i_n}\,.$$
Naturally, $\mathscr{J}_{n,k}\in\mathcal{B}_{n,k}$. The polymatching described in that tensor could not be more simple: the $k$ views $\left(\*x_i^1, \dots, \*x_i^k\right)$ of each point $i\leq n$ are matched together.

\textbf{From Embeddings to Cost Tensors. } We use a multiway cost function $c: \mathbb{R}^{\tiny{\overbrace{d\times\dots\times d}^{k\, \text{times}}}}\rightarrow \R$ to construct, using the information contained in $\mathscr{X}$, a cost \textit{tensor} that evaluates that multiway cost on \textit{all} $n^k$ possible combinations of points (for each of the $k$ views, choose one among $n$ available points). We call $\mathscr{M}_c$ the operator from $ \mathbb{R}^{k\times n \times d}$ to $\mathcal{T}_{n,k}$, defined as:
$$
\mathscr{M}_c(\mathscr{X})= [c\left(\*x_{i_1}^{1}, \;\cdots,\; \*x_{i_k}^{k}\right)]_{i_1,\dots, i_k}\,,
$$
where all indices $1\leq i_1, \dots, i_k\leq n$. The multiway cost function $c$ can be seen equivalently as a function from $\mathbb{R}^{n\times d}$ to $\mathbb{R}$. While several costs have been considered in the MM-OT literature, e.g. repulsive Coulomb costs in density functional theory~\citep{pass2015multi}), we use the simplest cost in our setting, quantified as the norm of the average of $k$ points on the sphere (whose norm is necessarily smaller than $1$), following insights from directional statistics~\citep{ley2017modern}. We define first the \textit{resultant} length of a set of a points on the sphere,
$$
R^2(\*z_1,\dots, \*z_k) := \| \tfrac{1}{k}\sum_\ell \*z_\ell\|^2 = 1 - 2\tfrac{k}{k-1}\sum_{\ell<m} \|\*z_\ell - \*z_m\|^2
$$
Note that the rightmost reformulation above, using pairwise distances, allows for a more efficient computation, provided in Algorithm~\ref{alg:tenscost}. The \textit{circular variance} can quantify dispersion for these $k$ points as:
\begin{equation}\label{eq:cv}
c_{\textrm{cv}}(\*z_1,\dots, \*z_k) = 1 - R^2(\*z_1,\dots, \*z_k)\,.
\end{equation}
We have also tested an alternative, the MLE variance parameter of the wrapped Gaussian given these $k$ points, a.k.a. the circular standard deviation,
\begin{equation}\label{eq:csd}
c_{\textrm{csd}}(\*z_1,\dots, \*z_k) = - \log(R^2(\*z_1,\dots, \*z_k))\,.
\end{equation}
We use $c_{cv}$ by default in all experiments, and only consider $c_{csd}$ in the ablation studies in \S~\ref{sec:ablations}.

\begin{algorithm}[tb]
\caption{Cost Tensor from Embeddings \small{$\mathscr{M}_{c_{\text{cv}}}(\mathscr{X})$}}\label{alg:tenscost}
\begin{algorithmic}
   \STATE {{\bfseries input:} embeddings tensor $\mathscr{X}\in\mathbb{R}^{k\times n \times d}$}
   \STATE {$\mathscr{A} = \mathbf{0} \in \mathcal{T}_{n,k}$}
   \STATE {$\mathscr{W} = \mathscr{X}(\texttt{None}, :, \texttt{None})\cdot \mathscr{X}(:, \texttt{None}, :, \texttt{None})$}
   \STATE {$\mathscr{D} = 2 - 2 \mathscr{W}.\mathrm{sum}(-1))$}
   \FOR{$\ell, 1\leq \ell\leq k$}
        \FOR{$m, \ell+1\leq m\leq k$}
            \STATE $I = \llbracket \ell-1 \rrbracket + \llbracket \ell+1, m-1 \rrbracket + \llbracket m+1, k \rrbracket$
            \STATE $\mathscr{A} = \mathscr{A} + \mathrm{expand\_dims}(\mathscr{D}[\ell,m,...], I-1)$
        \ENDFOR
   \ENDFOR
   \STATE {\bfseries output:} Cost tensor \small{$\mathscr{M}_{c_{\text{cv}}}(\mathscr{X})=\left(\tfrac{2}{k-1}\right)^2\mathscr{A}\in\mathcal{T}_{n,k}$}
   \vspace{-0.12mm}
\end{algorithmic}
\end{algorithm}

\subsection{Multi-Marginal Matching Gap}\label{subsec:MMMG}

With these definitions, we can define the $\mfg$ loss:
\begin{definition}\label{def:m3g} The multimodal multi-marginal matching gap (M3G) of data tensor $\mathscr{X}$, parameterized by a multiway cost $c$ and $\varepsilon>0$, is the gap to optimality of the ground-truth matching tensor $\mathscr{J}_{n,k}$:
\begin{align}\label{eq:mfg}
\!\!\!\!\mfg_{c,\varepsilon}&(\mathscr{X})\! :=\! h_{k,\varepsilon}(\mathscr{J}_{n,k}, \mathscr{M}_c(\mathscr{X})) - \!\!\inf_{\mathscr{P}\in\mathcal{B}_{n,k}}\!\!h_{k,\varepsilon}(\mathscr{P}, \mathscr{M}_c(\mathscr{X}))\nonumber\\
&\!\!\!\!\! = \langle \mathscr{J}_{n,k}, \mathscr{M}_c(\mathscr{X}) \rangle +\varepsilon\log n - \OT_{k,\varepsilon}(\mathscr{M}_c(\mathscr{X}))\,.
\end{align}
\end{definition}

The idea of contrasting the loss of a ground-truth solution to that achieved by a solver parameterized by actionable inputs (here, ultimately, the encoder parameters) can be traced back to, e.g. structured SVMs~\citep{JMLR:v6:tsochantaridis05a}, and was investigated in more depth in the elegant framework of Fenchel-Young losses~\citep{blondel2020learning}. In the context of OT, a similar idea was used to define a regularizer for vector-to-vector mappings~\citep{uscidda2023monge}.

\begin{proposition}\label{prop:m3g} The $\mfg$ loss is non-negative. The gradient of the $\mfg$ losses only requires applying the vector-Jacobian operator~\citep[\S 2.3.5]{blondel2024elements} of $\mathscr{M}$, $\partial\,\mathscr{M}(\cdot)^*[\cdot]$, evaluated at $\mathscr{X}$, to the difference of two polystochastic tensors, the ground-truth $\mathscr{J}_{n,k}$ and the optimal $\mathscr{P}_\varepsilon(\mathscr{M}(\mathscr{X}))$ given in~\eqref{eq:solregwassk}:
    $$\nabla\mfg(\mathscr{X}) = \partial\,\mathscr{M}\left(\mathscr{X}\right)^*\left[ \mathscr{J}_{n,k} - \mathscr{P}_\varepsilon(\mathscr{M}(\mathscr{X}))\right]\in\mathbb{R}^{k\times n \times d}\,.$$
\end{proposition}

These results come from an application 
of Fenchel-Young losses~\citep{blondel2020learning}. Briefly, the first result comes from the fact that $\mfg$ is an optimality gap; the second follows from the fact that $\OT_{k,\varepsilon}$ is an unconstrained convex optimization problem, and therefore an application of \citeauthor{danskin2012theory}'s theorem (assuming Alg.~\ref{alg:mms} is run to low tolerance $\alpha$, which we do by setting it to $10^{-3}$) states that $\nabla\OT_{k,\varepsilon}(\mathscr{C})=\mathscr{P}_\varepsilon(\mathscr{C})$. This, combined with the chain-rule, gives the result.

\textbf{Deeper Dive into $k=2$.}
Although the case $k=2$ is not the main focus of our work, we highlight that $\mfg$ does not reduce to an InfoNCE-like loss, even for two views. %
Indeed, for $k=2$ only, and using notations from \S~\ref{sec:back_infonce} one recovers, up to the constant $\varepsilon \log n$, that:
$$\!\!\mfg_{c,\varepsilon}([\*X^1\!,\*X^2])=\! \tfrac1n\|\*X^1\!-\*X^2\|^2 - 
 \OT_{2,\varepsilon}\left(\!\left[c(\*x^1_i,\*x^2_j)\right]_{ij}\!\right).$$
For $k=2$, the $\mfg$ loss provides an alternative to the classic InfoNCE loss, and is related, but not equivalent to, the recent ``inverse optimal transport'' approach advocated in~\citep{shi2023understanding}. We briefly discuss this link in \S\ref{sec:related}.

\begin{table*}[htb!]
\caption{\textbf{Multiview models performance as a function of number of views, $k$, for models pre-trained on ImageNet-1k.} 
    Evaluation of classification performance of $\mfg$ ($\varepsilon=0.2$) in comparison to paiwise losses, $\rm{BYOL}$ and $\rm{InfoNCE}$  extended to multiview using either the \textit{pairwise} sum across views ($\mathcal{L}_{\rm{pwe}}$), or a \textit{one vs. average-of-rest} ($\mathcal{L}_{\rm{ave}}$). We evaluate the performance for varying $k$ views, with $n=64$ batch size, trained for $300$ epochs. Reported are mean and standard variation over five independent repetitions per setting. In bold is the top performing method per setting.}
\label{tab:inet}
\vskip 0.15in
\begin{center}
\begin{small}
    \begin{tabular}{@{} l c c c c c @{}}
      \toprule
\multirow{2}{*}{\# Views} & \multicolumn{5}{c}{Method}  \\   
\cmidrule{2-6}
& $\rm{BYOL}_{\rm{ave}}$  & $\rm{BYOL}_{\rm{pwe}}$ & $\rm{InfoNCE}_{\rm{ave}}$ & $\rm{InfoNCE}_{\rm{pwe}}$  & $\mfg$ \\
       \midrule    
     $k=2$ & \multicolumn{2}{c}{$74.62 \pm 0.14$} &  \multicolumn{2}{c}{$74.61 \pm 0.16$} & $\bf{74.75 \pm 0.48}$    \\
       $k=3$ &  $74.60 \pm 0.16$  &  $75.16 \pm 0.09$  &  $74.24 \pm 0.13$ & $75.36 \pm0.10$ & $\bf{75.61 \pm 0.12}$  \\
       $k=4$ &  $75.04 \pm 0.18$  & $75.06 \pm 0.10$ & $74.80 \pm 0.09$  & $75.26 \pm 0.16$ & $\bf{75.75 \pm 0.11}$  \\
      \bottomrule
 \end{tabular}
\end{small}
\end{center}
\vskip -0.1in
\end{table*}

\subsection{Learning Representations with $\mfg$}
Suppose we are given a batch of $n$ objects $z_{1}, \dots, z_{n}$, and that each of these objects is available in $k$ multiple views, either through data collection or augmentations $((z_{i}^{1}, \dots, z_{i}^{k}))_i$. 
Broadly speaking, we consider parameterized networks, $f_{\theta_\ell}$, $\ell\leq k$, in which case $\theta$ would stand for the list of all parameters $(\theta_\ell)_\ell$. We assume that all networks take values in the $d$-sphere, $\{\*x\in\mathbb{R} : \|\*x\|=1\}$. 
We propose to minimize the $\mfg$ loss on the $n\times k$ encodings of all these objects, for each minibatch.
$$
\mathcal{L}(\theta) := \mfg_{c,\varepsilon}\left( [f_{\theta_\ell}(z_i^\ell)]_{\ell,i}\right)\,.
$$

\section{Experiments}\label{sec:experiments}
We test the $\mfg$ loss in an SSL %
setting (ImageNet-1k) and two multimodal tasks (DomainNet and PhysioNet). 
In \S~\ref{sec:imagenet} and \S~\ref{sec:donmainnet}, we use a joint embedding student-teacher architecture, see \S~\ref{sec:back_student_teacher}. The student network is evaluated on all modalities $[1,k]-i$, apart from one index $1\leq i\leq k$, for which the teacher is used. Gradients are aggregated on the $k-1$ evaluations. Index $i$ loops then across all $k$ modalities to form an aggregated loss. 
In\S~\ref{sec:eeg} we use a single common network, as proposed by~\citep{brusch2023multi}.

\subsection{Multiview SSL Performance on ImageNet-1k}\label{sec:imagenet}
We use $k$ random augmentations for each image, and study the impact of $k$ on the linear performance of encoder models pre-trained on ImageNet-1k~\citep{deng2009imagenet}. We compare our loss $\mfg$, with the previously suggested extensions of contrastive losses to $k \geq 3$, using either aggregation of \textit{pairwise} contributions  ($\mathcal{L}_{\rm{pwe}}$, \eqref{eq:sum}), or the \textit{one vs. average-of-rest} approach ($\mathcal{L}_{\rm{ave}}$, \eqref{eq:mvp}). %
We train and evaluate each setting in five independent repetitions, and report mean and standard deviation. 

\textbf{Augmentations.} We use the augmentations introduced in BYOL~\citep{grill2020bootstrap} and SimCLR~\citep{chen2020simple}. These vary in the parameters used for the aggregated transformations--cropping, random flipping, random color jittering, Gaussian blur, and grayscale or solarization. See \appref{app:impl_details} for details on the $k$ augmentation stacks. 

\textbf{Architecture.} The backbone encoder is a ViT-B/16 architecture~\citep{dosovitskiy2020image}, followed by an MLP projection head (see \appref{app:impl_details} for details). We use a batch size of $n=64$ per GPU. We train all models for $300$ epochs.

\textbf{Results.} Table~\ref{tab:inet} indicates that the $\mfg$ loss performs slightly better than baselines alternatives for multiview learning, improving the linear classification accuracy by $.25\%$ and $.49\%$ respectively for $k=3$ and $k=4$, validating the soundness of $\mfg$ with a fairly small batch size.

\textbf{On Increasing Batch size.} As mentioned earlier in this section, results reported here use $k-1$ student branches vs. $1$ teacher. This allows dropping the forward activations of the teacher branch. Informally, for a batch size of $n$, given
$s$ student branches,
$k-s$ teacher branches, and writing
$M$ for the memory cost needed to store all activations for a parameter $\theta$ (of a single student branch, for a single point)
this yields a total memory cost of $\approx O(s n M+ n^k)$, taking into account the cost of the MM-S cost tensor. This results in a trade-off, since smaller $s$ can allow for larger $n$ of $k$, depending on the magnitude of $M$. We leave this direction for future research.

\subsection{Multimodal Domain Adaptation} \label{sec:donmainnet}

\begin{table*}[htb!]
\caption{\textbf{Predictions accuracy over unseen unlabeled domains over the DomainNet dataset.} 
    Evaluation of classification performance of $\mfg$ ($\varepsilon=0.05$) compared to two pairwise losses, considering $\mathcal{L}_{\rm{pair}} \in \left\{ \rm{InfoNCE}, \rm{BYOL}\right\}$. For each we evaluate the \textit{pairwise} sum across views ($\mathcal{L}_{\rm{pwe}}$), and a \textit{one vs. average-of-rest} ($\mathcal{L}_{\rm{ave}}$). All models are trained with a fixed batch size of $n=16$ for 300 epochs. Mean and standard deviation of performance reported for four independent repetitions. In bold is the top performing method per setting.}
\label{tab:domainnet}
\vskip 0.15in
\begin{center}
\begin{small}
    \begin{tabular}{@{} c c c c c c @{}}
      \toprule
Domains & \multicolumn{5}{c}{Method}  \\   
 \cmidrule{2-6}
seen $\rightarrow$ unseen  & $\rm{BYOL}_{\rm{ave}}$  & $\rm{BYOL}_{\rm{pwe}}$ & $\rm{InfoNCE}_{\rm{ave}}$ & $\rm{InfoNCE}_{\rm{pwe}}$  & $\mfg$  \\
       \midrule   
{\it{$\lnot$clp $\rightarrow$ clp}}  & $24.1 \pm 0.2$ & $9.6 \pm 4.8$  & $23.2 \pm 0.4$ & $6.9 \pm 10.3$ & $\bf{32.4 \pm 0.3}$  \\ 
 {\it{$\lnot$inf $\rightarrow$ inf}} & $10.2 \pm 0.1$ & $5.8 \pm 0.8$  & $11.0 \pm 0.2$  & $10.1 \pm 0.4$ &  $\bf{12.2 \pm 0.1}$ \\
 {\it{$\lnot$pnt $\rightarrow$ pnt}} & $28.3 \pm 0.1$ & $21.3 \pm 0.9$ & $30.6 \pm 0.6$  & $24.7 \pm 1.8$ &  $\bf{31.3 \pm 0.1}$  \\
 {\it{$\lnot$qdr $\rightarrow$ qdr}} & $7.8 \pm 0.2$  & $3.4 \pm 2.2$  & $8.8 \pm 0.2$ & $8.5 \pm 0.5$ &  $\bf{10.4 \pm 0.3}$  \\   
 {\it{$\lnot$rel $\rightarrow$ rel}} & $43.1 \pm 0.3$ & $30.9 \pm 9.3$  & $44.6 \pm 0.1$ & $42.0 \pm 0.6$ &  $\bf{46.3 \pm 0.2}$ \\
 {\it{$\lnot$skt $\rightarrow$ skt}} & $21.6 \pm 0.6$  & $10.1 \pm 2.1$  & $24.9 \pm 0.1$& $20.5 \pm 1.6$   &  $\bf{26.5 \pm 0.4}$ \\
      \bottomrule
\end{tabular}
\end{small}
\end{center}
\vskip -0.1in
\end{table*}

\begin{figure*}[htb!]
    \centering
    \includegraphics[width=\textwidth]{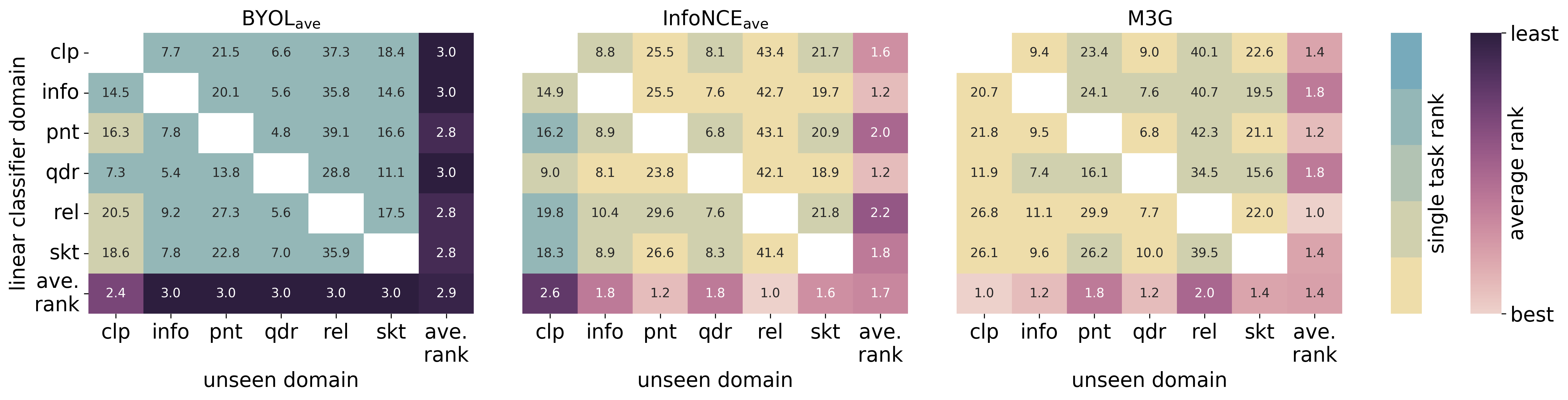}
    
\caption{\textbf{Pairwise domain prediction accuracy on the DomainNet dataset.} 
    The prediction accuracy over the unseen domain using a linear classifier trained on a single domain in the pre-training train set. Each table presents the performance of a different model choice. From left to right, baseline approaches, using the pairwise losses to evaluate \textit{one vs. average-of-rest},  
    $\rm{BYOL}_{\rm{ave}}$ (left)  and $\rm{InfoNCE}_{\rm{ave}}$ (center), compared to 
    $\mfg$ (right). Columns correspond to the unseen domains and rows to the domains used for the linear classifier training. Mean performance reported for four independent repetitions. }
\label{fig:domainnet_pairs}
\end{figure*}

We consider a domain adaptation (DA) task, where the goal is to learn a common encoder, followed by one or multiple classifiers, using labeled data from multiple domains. We quantify the generalization power of this pre-trained encoder with a classification task, tested on data coming from a new, completely unseen domain~\citep{peng2019moment,gulrajani2020search}. We assume that the new data, despite coming from an unseen domain, still falls in the same classes. For that purpose, we use the DomainNet dataset~\citep{peng2019moment}. The dataset consists of 569,010 images divided into 345 different categories, and subdivided in 6 different domains--real photos (\textit{rel}), cliparts (\textit{clp}), sketches (\textit{skt}), infographic images (\textit{inf}), artistic paintings (\textit{pnt}), and quickdraw (\textit{qdr}).

\textbf{Training Procedure.} We consider the same losses: $\mfg$ and the baselines, $\rm{InfoNCE}$ and $\rm{BYOL}$ (each evaluated using both aggregation choices, $\rm{pwe}$ and $\rm{ave}$). We pick one domain that acts as the unseen modality, and train representations on the $k=5$ remaining domains. Following the conventions set in \S\ref{sec:methods}, the $n\times k$ points are sampled by picking randomly $n$ classes, and for each class, $k$ images in the dataset, coming from each of the $k=5$ domains. For each of the settings we pre-train four independent representation encoders. The encoders' backbone is identical to that used on the ImageNet-1 dataset (ViT-B-16). We repeat this for each of the six modalities--implying a total of $5\times 4 \times 6 =120$ models. All models are trained using the same architecture and parameters choice (for details see \appref{app:impl_details}), and evaluated on two tasks:
 
\textbf{5 Domains \textit{vs}. 1.} For the first assessment, we train a linear classifier jointly on all 5 seen domains, and report test results on the unseen domain, see Table~\ref{tab:domainnet}. While prediction accuracy in this task varies according to domain, ranging from $~11\%$ to $~30\%$ for $\mfg$, $\mfg$ is consistently ranked $1$st, with a mean improvement of $3.1\%$ over the $2$nd best model ($\rm{InfoNCE}_{\rm{ave}}$). In contrast to the ImageNet-1k task (\S\ref{sec:imagenet}), the \textit{ave} baseline outperforms the \textit{one vs. rest}.

\textbf{1 Domain \textit{vs.} 1.} Next, we consider a harder task, training five independent classifiers, one per domain. Figure~\ref{fig:domainnet_pairs} reports the prediction accuracy of these classifiers when tested using images from the unseen domain. The performance of the \text{pwe} approach is overall much worse in this task as well, and presented in \S~\ref{app:results}, 
 Figure~\ref{fig:domainnet_pairs_all}.

We find that globally $\mfg$ outperforms baseline approaches, with an average rank of $1.4$ across all tasks ($30$ linear evaluations $\times$ 4 independent repetitions $=120$).

\subsection{EEG Data}\label{sec:eeg}
Health records often contain multi-channel time series data, available in vast amounts, but that require manual annotations by domain experts. Because channels provide aligned data points, they provide a testbed for multimodal embedding approaches. We apply directly the $\mfg$ loss on an EEG dataset using $k=6$ channels, taken from the PhysioNet Challenge 2018~\citep{goldberger2000physiobank,ghassemi2018you}. EEG is a neurophysiological technique that records and measures the brain’s electrical
activity. The train data contains segmented samples of $994$ individuals, and the evaluation dataset, SleepEDFx~\citep{goldberger2000physiobank,kemp2000analysis}, contains $153$ nights of sleep recordings from $78$ individuals, each annotated as belonging to one among five classes of sleep stage. 

\textbf{Classification.} The task is to accurately predict the sleep stage using a sample of $30$s. Freezing the pre-trained representation model, we train a linear encoder over samples of $10$, $50$, $100$, and $1000$ data points for each of the five classes, and evaluate the prediction accuracy over the same number of samples respectively. We reuse the codebase provided by~\citep{brusch2023multi}.

\textbf{Results.} In Table~\ref{tab:ts_ssl} we compare $\mfg$ to the \textit{pairwise} InfoNCE loss. In accordance with previous results we find the $\mfg$ performs better than $\rm{InfoNCE}_{\rm{pwe}}$. 

\begin{table}[htb!]
\caption{\textbf{Prediction accuracy over multichannel EEG dataset.} 
    Linear prediction accuracy of $\mfg$ and baseline method on a $k=6$ EEG channels dataset using different numbers of samples per class $s \in \{ 10, 50, 100, 1000\}$. We report mean results, averaged over 5 seeds. $\rm{InfoNCE}_{\rm{pwe}}$ models are trained using a batch size of $n=64$ and $\mfg$ uses a batch size of $n=16$. All models are pre-trained for $10$ epochs and fine-tuned on the classification task for a maximum of $40$ epochs.}
\label{tab:ts_ssl}
\vskip 0.15in
\begin{center}
\begin{small}
    \begin{tabular}{@{} l c c c c @{}}
      \toprule
 \multirow{2}{*}{Method} & \multicolumn{4}{c}{samples per class} \\   
 \cmidrule{2-5}
       & $s=10$ & $s=50$ & $s=100$ & $s=1000$\\
       \midrule    
       $\rm{InfoNCE}_{\rm{pwe}}$ & $35.5$ & $\textbf{49.3}$ & $52.2$ & $56.3$ \\
       $\mfg, \varepsilon=0.2$ & $\textbf{36.6}$ & $49.2$ & $\textbf{56.1}$&  $\textbf{64.6}$  \\
      \bottomrule
 \end{tabular}
\end{small}
\end{center}
\vskip -0.1in
\end{table}

\section{Ablation studies}\label{sec:ablations}

\begin{figure}[htb!] 
\includegraphics[width=0.45\textwidth]{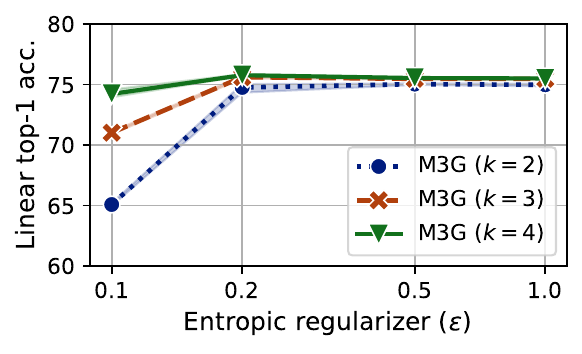}
\caption{\textbf{Linear performance on ImageNet-1k as a function of the entropic regularizer $\varepsilon$.} We report the linear top-1 accuracy for different values of the MM-OT entropic regularizer $\varepsilon$, as we vary the view multiplicity, $k$. All results are given for the same batch size ($n=64$) and training duration ($300$ epochs). Solid line and band depict the mean and $95\%$ confidence interval over five independent repetitions.}
\label{fig:entropic}
\end{figure}

The $\mfg$ loss is parameterized by a multiway cost function $c$, and by entropic regularization $\varepsilon$. We study how these two choices affect performance. We limit our study to the ImageNet-1K multiview task.

\textbf{Entropic Regularization.} Because embeddings are always normalized, and costs depend directly on dot-products, the range of cost values is constrained. Thanks to this, setting $\varepsilon$ was fairly easy. As shown in Fig.~\ref{fig:entropic}, and observed in most of our other experiments, overall performance is fairly robust to $\varepsilon$. Setting $\varepsilon=0.2$ returned consistently good results.

\textbf{Cost Function.} The multiway cost function $c$ is the other important degree of freedom available to the user to shape the $\mfg$ loss. Apart from the circular variance used throughout our experiments ($\mfg_{\rm{cv}}$) we evaluate the performance using a circular standard deviation cost ($\mfg_{\rm{csd}}$). We observe that $\mfg$ is robust to this choice, attaining similar performance under both cost choices, see Table~\ref{tab:cost_ablation}.

\begin{SCtable}[][htb!]
\caption{\textbf{Robustness to cost function.} 
    Classification performance of $\mfg$ models pre-trained on ImageNet-1k, using either $c_{\rm{cv}}$,~\eqref{eq:cv} or $c_{\rm{csd}}$,~\eqref{eq:csd}, with $\varepsilon=0.2$. 
    }
\label{tab:cost_ablation}
\begin{small}
    \begin{tabular}{@{} l c c  @{}}
    \\
      \toprule
\multirow{2}{*}{Views} & \multicolumn{2}{c}{Method}  \\   
\cmidrule{2-3}
 & $\mfg_{\rm{cv}}$ & $\mfg_{\rm{csd}}$ \\
       \midrule    
     $k=2$ &  $74.75$   & $73.81$ \\
       $k=3$ &   $75.61$  & $75.63$  \\
       $k=4$ &   $75.75$  &  $75.73$  \\
      \bottomrule
 \end{tabular}
\end{small}
\end{SCtable}

\begin{figure}[htb] 
\includegraphics[width=0.45\textwidth]{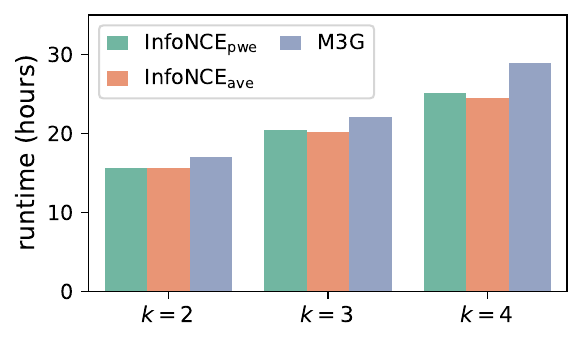}
\vspace{-.2in}
\caption{\textbf{Compute overhead incurred by $\mfg$ on ImageNet-1k as a function of $k$.}  All results are given for the same per GPU batch size ($n=64$), $300$ epochs, $\varepsilon=0.2$ for $\mfg$, run on 4 nodes of 8 A100 GPUs.}
\label{fig:iterations}
\end{figure}

\textbf{Compute Overhead.} Despite the daunting cost of running the MM-\citet{sinkhorn1964relationship} algorithm in $n^k$, we show that in the most computationally demanding of our tasks (ImageNet-1k), using $\mfg$ only incurs a relatively minor compute overhead. This is summarized in Figure~\ref{fig:iterations}.

\section{Discussion}
\subsection{Related Works}\label{sec:related}
\textbf{MM-OT.} When the number of views $k\geq 3$, we are not aware of any other work that uses MM-OT to study multiple representations of objects. Compared to regular OT, MM-OT has been used in far fewer applications, notably to handle density functional theory in chemistry with Coulomb costs~\citep{pass2015multi,benamou2017numerical}. MM-OT has very recently started playing a role in core ML tasks, e.g. with recent links to adversarial multiclass classification \citep{trillos2023multimarginal}. Much like regular OT, MM-OT  has also been extended to accommodate unbalanced constraints~\citep{beier2023unbalanced} or quadratic (Gromov-Wasserstein-like) objectives~\citep{beier2022multi}. Solving the MM-OT problem raises many challenges that are increasingly better understood in theory~\citep{pmlr-v151-le22a,lin2022complexity,altschuler2021hardness}. Finding alternative schemes to compute or approximate MM-OT is a very recent and active research subject, using e.g. ODEs~\citep{nenna2023ode} or by exploiting a more specific structure in costs~\citep{haasler2021multi,haasler2021multimarginal}. These ideas might be employed to speed up our scheme, as using \citeauthor{danskin2012theory}'s theorem leaves ample room for solving MM-OT with any forward pass, without having to go through a differentiable solver.

\textbf{The case \textbf{$k = 2$}.} When restricting our contribution to the simplest case $k=2$, our method reduces to a ``classic'' OT formulation, solved with the usual Sinkhorn algorithm. Closest to our method lies the proposal of~\citet{shi2023understanding} to use Inverse OT as a loss in SSL. There is, however, a significant discrepancy between their work and our loss for $k=2$: ~\citet{shi2023understanding} define their loss as the KL divergence between the ground truth identity matching matrix $\*J_{n}$, and the optimal coupling returned by Sinkhorn $\*P_\varepsilon(\*C(\*X))$. To compute the gradient of their loss, they need, therefore, to differentiate through Sinkhorn iterations. While this can be done by either unrolling iterations, or using the implicit function theorem~\citep{luise2018differential,cuturi2022optimal}, this adds memory and compute requirements. Because $\mfg$ is an optimality gap, namely a Fenchel-Young loss~\citep{blondel2020learning}, we can avoid that backward pass thanks to \citeauthor{danskin2012theory}'s theorem. The recent proposal of~\citet{jiang2023hard} is also closely related to $\mfg$ loss when $k=2$, since they propose to apply OT weights within the negative sample reweighting approach of~\citet{robinson2020contrastive}.

\subsection{Limitations}
An important limitation of the $\mfg$ loss lies in solving an MM-OT problem, using the MM-\citeauthor{sinkhorn1964relationship}, see Alg.~\ref{alg:mms}. Computing the $\mfg$ loss incurs a cost that scales as $n^k$, preventing, in practice, using large batch sizes. We believe this exponential scaling is likely the price to pay to account simultaneously for all $k$-tuples of views. Our experiments show that for small batch sizes (e.g. $n=16,32,64$) and small $k$ (we considered $k\leq 6$), this compute overhead was reasonable, notably when paired with large encoders, such as ViT-B/16 models (See~\ref{fig:iterations}). However, this increase becomes intractable for larger $k$ values if one uses, as we did, a generic multiway cost. Our loss is, however, fairly simple, since it only has two hyperparameters: a multiway cost $c$ and the $\varepsilon$ regularization. We have observed good performance for most $\varepsilon$ choices but suspect that $\varepsilon$ should scale with the batch-size for best performance. We studied an alternative cost ($\rm{csd}$), but this did not yield significantly different results. As future work, we believe one might explore better cost functions, either for computational or modeling reasons, e.g. using domain-specific knowledge that focuses on specific subsets of the $k$ views.

\subsection{Conclusion}
To our knowledge, the $\mfg$ loss is the first contrastive loss proposed to learn multi-representations that takes a holistic view of all $k$ views (when $k\geq 3$). Specifically, $\mfg$ avoids falling back to contrasting views in a pairwise approach. The $\mfg$ loss scores the coherence of a set of $n\times k$ point embeddings, by contrasting the polymatching cost of a known ground-truth matching with the best polymatching that can be obtained, as approximated by the multi-marginal~\citeauthor{sinkhorn1964relationship} algorithm. The $\mfg$ loss and its gradient can be computed with a single forward execution of the multi-marginal~\citeauthor{sinkhorn1964relationship} algorithm. While the application of the $\mfg$ loss to practical tasks may seem daunting, because of the exponential complexity in $k$ incurred when running the MM-\citeauthor{sinkhorn1964relationship} algorithm, we show that the overhead paid to compute this loss, in terms of running time, is manageable as long as batch size $n$ and $k$ are not too large. We have presented promising performance on a variety of self-supervised and multimodal tasks, paving the way for future extensions that can leverage more informed cost structures.

\subsection{Broader Impact}
This paper presents work whose goal is to advance the field of Machine Learning. There are many potential societal consequences of our work, none which we feel must be specifically highlighted here.

\bibliography{main}
\bibliographystyle{icml2024}

\clearpage
\appendix

\setcounter{table}{0}
\setcounter{figure}{0}
\renewcommand{\thetable}{A\arabic{table}}
\renewcommand{\thefigure}{A\arabic{figure}}

\section{Implementation Details} \label{app:impl_details}

\paragraph{Multi-Marginal Sinkhorn Optimization.}
To perform experiments, we implemented the multi-marginal Sinkhorn algorithm (Alg.~\ref{alg:mms})  in PyTorch~\citet{paszke2019pytorch}. 

\paragraph{Hyperparameters for Models Training.}
In Table~\ref{tab:mg_recipe} we provide the hyperparameters used to train ImageNet-1k and DomainNet models. In all cases the encoder is based on ViT-B/16 architecture and the following projection and predictor heads consist of a linear layer with output size $4096$ followed by Gaussian error linear units (GeLU)~\citep{hendrycks2016gaussian}), and an additional linear layer with output dimension $256$.

For the EEG dataset, we follow the setting reported in~\citep{brusch2023multi}, using the implementation provided in the GitHub repository\footnote{\href{https://github.com/theabrusch/Multiview_TS_SSL}{https://github.com/theabrusch/Multiview\_TS\_SSL}}. The network is composed of six convolutional blocks consisting of a 1D convolution, a dropout layer, a group normalization layer, and a GELU activation function. The kernel width and stride is three in the first layer and two in the remaining five layers. $256$ kernels are used for all intermediate layers and the final output dimension is $64$. A readout layer with kernel width and stride set to 1 is added at the end. We train models for 5 different seeds and report avergage results. All models are trained for $10$ epochs and a batch size of $n=16, 64$ for $\mfg, \rm{InfoNCE}_{\rm{pwe}}$ respectively.  

\paragraph{Augmentations.}
For image datasets (ImageNet-1k and DomainNet), we use  augmentation settings introduced in BYOL~\citep{grill2020bootstrap} and SimCLR~\citep{chen2020simple}. We provide the pseudocode for the augmentations used in Pseudocode~\ref{lst:augs}. For ImageNet-1k the maximal stack ($k=4$) is defined as \texttt{A = [byol-global1, byol-global2, simclr, byol-global1]}, for lower $k$ we take \texttt{A[:k]}. For DomainNet training we use the \texttt{simclr} augmentation for all views. In all cases, for test augmentations we follow the standard practice--resize, center crop and normalization.

\lstset{
  breaklines=true,
  basicstyle=\fontsize{8pt}{8pt}\ttfamily\selectfont,
} \label{alg:augs}
\begin{lstlisting}[language=python, caption={Definition of the train augmentations.}, label={lst:augs}]
    byol-global1 = [
        RandomResizedCrop(
            size=224, 
            scale=(0.08, 1.0), 
            interpolation=Image.BICUBIC
        ),
        RandomHorizontalFlip(p=0.5),
        RandomApply([
                ColorJitter(
                    brightness=0.4, 
                    contrast=0.4, 
                    saturation=0.2,
                    hue=0.1
                )
            ], p=0.8,
        ),
        RandomGrayscale(p=0.2),
        GaussianBlur(),  
        Normalize(
            mean=(0.485, 0.456, 0.406), 
            std=(0.229, 0.224, 0.225)
        )
    ]
    
    byol-global2 = [
        RandomResizedCrop(
            size=224, 
            scale=(0.08, 1.0), 
            interpolation=Image.BICUBIC
        ),
        RandomHorizontalFlip(p=0.5),
        RandomApply([
                ColorJitter(
                    brightness=0.4, 
                    contrast=0.4, 
                    saturation=0.2,
                    hue=0.1
                )
            ], p=0.8,
        ),
        RandomGrayscale(p=0.2),
        RandomApply([GaussianBlur()], p=0.1),
        RandomApply([Solarization()], p=0.2),
        Normalize(
            mean=(0.485, 0.456, 0.406), 
            std=(0.229, 0.224, 0.225)
        )
    ]
    
    simclr = [
        RandomResizedCrop(
                size=224, 
                scale=(0.08, 1.0), 
                interpolation=Image.BICUBIC
            ),
        RandomHorizontalFlip(p=0.5),
        RandomApply([
                ColorJitter(
                    brightness=0.8, 
                    contrast=0.8, 
                    saturation=0.8,
                    hue=0.2
                )
            ], p=0.8,
        ),
        RandomGrayscale(p=0.2),
        RandomApply([GaussianBlur(kernel_size=23, sigma=[0.1, 2.0])], p=0.5),
        Normalize(
            mean=(0.485, 0.456, 0.406), 
            std=(0.229, 0.224, 0.225)
        )
    ]
\end{lstlisting}

\begin{table}[htb!]
  \caption{\textbf{Vision models hyperparameters.}}
  \label{tab:mg_recipe}
  \centering
  \small
  \begin{tabular}{lc}
    \toprule
    Encoder architecture & ViT-B/16 \\
    Weight initialization & \texttt{trunc\_normal(.02)}  \\
    Backbone normalization    & LayerNorm  \\
    \multirow{2}{*}{Batch size} & 2048 (ImageNet-1k)  \\
                             & 512 (DomainNet)  \\
    Head normalization    & LayerNorm  \\
    Synchronized BatchNorm over replicas & True \\     
    Learning rate schedule & Single Cycle Cosine \\    
    Learning rate warmup (epochs) & 10 \\    
    Learning rate minimum value & $5\times 10^{-5}$  \\    
    Training duration (epochs) & 300 \\
    Optimizer & AdamW \\    
    Optimizer scaling rule & Adam \\
    Base ($\beta_1, \beta_2$) & (0.9, 0.95) \\
    Base learning rate & $6.5 \times 10^{-4}$  \\
    \multirow{2}{*}{Per GPU Batch size} & 64 (ImageNet-1k)  \\
                             & 16 (DomainNet)  \\
    Base teacher momentum & 0.99 \\    
    Weight decay & 0.04 \\
    Weight decay end & 0.4 \\
    Weight decay warmup & 0.0 \\ 
    \bottomrule
  \end{tabular}
\end{table}

\section{Training and Evaluation} \label{app:evaluation}

\paragraph{Linear Evaluation of Image Models.}
For the image datasets tasks (ImageNet-1k and DomainNet) we follow the standard linear evaluation pipeline~\citep{chen2021empirical, he2016deep}. We freeze the backbone encoder of the pre-trained model and train a linear classifier for $100$ epochs on the data used for pre-training (ImageNet-1k or DomainNet respectively). We use the SGD optimizer with zero \texttt{weight\_decay}. For the \texttt{learning\_rate} we sweep over two possible values ($0.01, 0.001$). Random horizontal flipping, random resized cropping and normalization are applied during training. 

\paragraph{DomainNet Downstream Evaluation}
Using a pre-trained model with an unseen domain we freeze the encoder and test the model performance in two regimes: (i) train a single linear classifier, as described above, using the five seen domains. (ii) train five different classifiers, each over a single seen domain. All linear classifiers are tested over the unseen domain.

\paragraph{Linear Evaluation of the EEG dataset.}
We follow precisely the evaluation suggested by~\citep{brusch2023multi}. Given the pre-trained model, consisting of a single encoder, a linear layer is used to combine all channel representations. A linear classifier is trained over the frozen joint representation for the classification task. Both layers are retrained from scratch.

\section{Additional Results} \label{app:results}

\paragraph{Pairwise domain evaluation over DomainNet dataset.}

\begin{figure*}[ht]
 \centering
    \includegraphics[width=\textwidth]{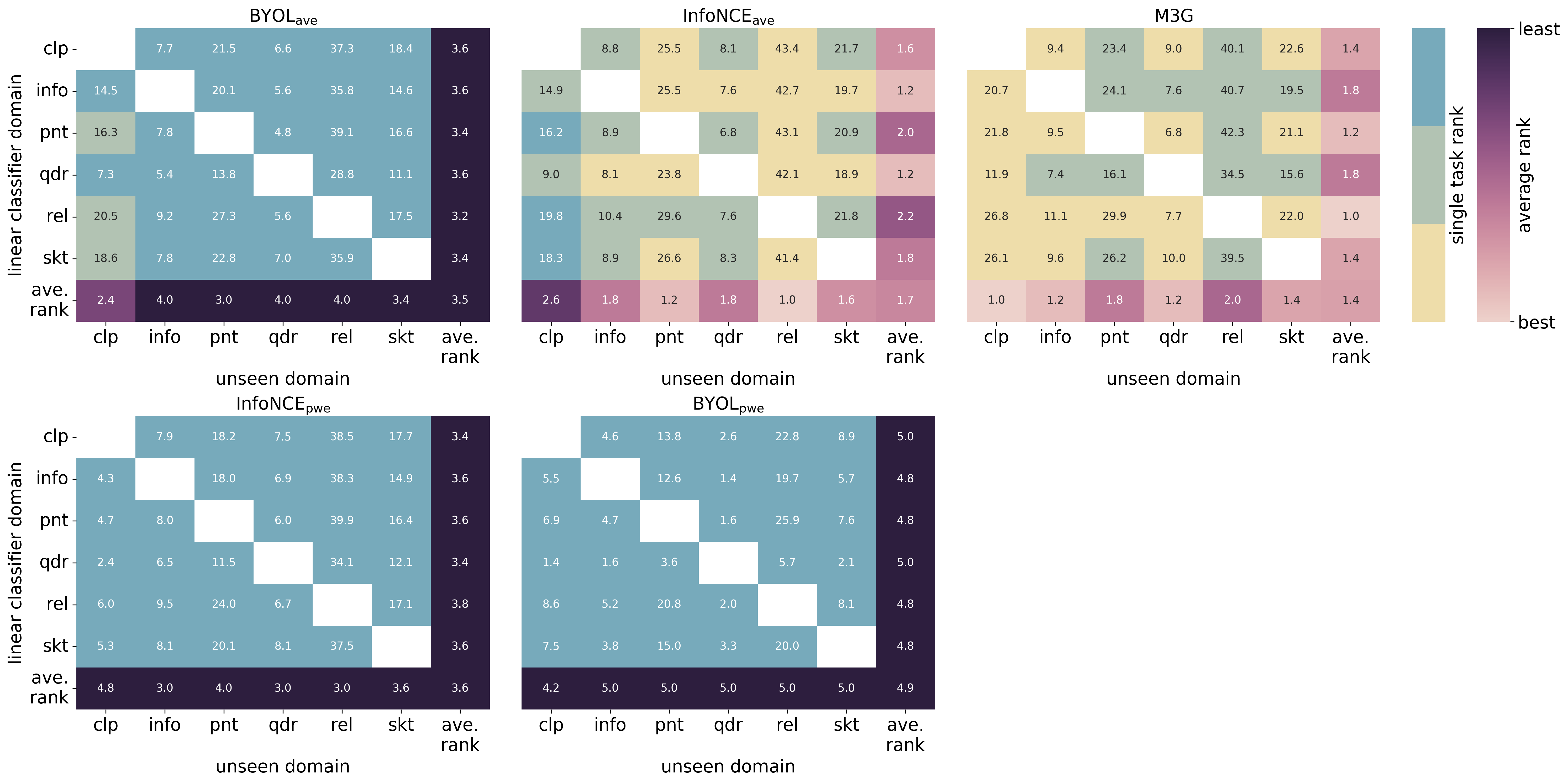}
    \caption{\textbf{Pairwise domain prediction accuracy on the DomainNet dataset.} 
     The prediction accuracy over the unseen domain using a linear classifier trained on a single domain in the pre-training train set. Each table presents the performance of a different model choice. Two left columns present baseline approaches, using the pairwise losses to evaluate \textit{one vs. average-of-rest} (top row) and \textit{pairwise} (bottom row). Columns from left to right consider 
    $\rm{BYOL}$ (left), $\rm{InfoNCE}$ (center), and
    $\mfg$ loss. 
    In each subplot, columns correspond to the unseen domains and rows to the domains used for the linear classifier training. Mean performance reported for four independent repetitions. }
    \label{fig:domainnet_pairs_all}
\end{figure*}

\end{document}